\documentclass[conference]{IEEEtran}
\IEEEoverridecommandlockouts
\usepackage{graphicx,mathtools,mathrsfs,amssymb}
\usepackage{multirow}
\usepackage{txfonts}
\usepackage{stfloats}
\usepackage{booktabs}
\usepackage{bm,url}

\usepackage[justification=centering]{caption}
\usepackage{multicol}
\usepackage{lipsum}   
\usepackage{titlesec} 
\usepackage{textcomp}
\usepackage{xcolor}
\def\BibTeX{{\rm B\kern-.05em{\sc i\kern-.025em b}\kern-.08em
    T\kern-.1667em\lower.7ex\hbox{E}\kern-.125emX}}
\begin{document}

\title{CIM-NET: A Video Denoising Deep Neural Network Model Optimized for Computing-in-Memory Architectures
\thanks{This work is funded by China Mobile Communications Group Co.,Ltd.}
}

\author{
\makebox[0.5\textwidth][c]{%
  \parbox{0.52\textwidth}{
    \centering
    \textbf{1\textsuperscript{st} Shan Gao} \\
    \textit{ZGC Institute of Ubiquitous-X Innovation and Applications} \\
    \textit{China Mobile Research Institute} \\
    Beijing, China \\
    pkugaoshan@pku.edu.cn
  }
}
\makebox[0.5\textwidth][c]{%
  \parbox{0.48\textwidth}{
    \centering
    \textbf{2\textsuperscript{nd} Zhiqiang Wu} \\
    \textit{China Mobile Research Institute} \\
    Beijing, China \\
    zhiqiang.wu@wright.edu
  }
}
\\[1.2em]  

\makebox[0.33\textwidth][c]{%
  \parbox{0.30\textwidth}{
    \centering
    \textbf{3\textsuperscript{rd} Yawen Niu} \\
    \textit{China Mobile Research Institute} \\
    Beijing, China \\
    niuyawen@chinamobile.com
  }
}
\makebox[0.33\textwidth][c]{%
  \parbox{0.30\textwidth}{
    \centering
    \textbf{4\textsuperscript{th} Xiaotao Li} \\
    \textit{China Mobile Research Institute} \\
    Beijing, China \\
    lixiaotao@chinamobile.com
  }
}
\makebox[0.33\textwidth][c]{%
  \parbox{0.30\textwidth}{
    \centering
    \textbf{5\textsuperscript{th} Qingqing Xu} \\
    \textit{China Mobile Research Institute} \\
    Beijing, China \\
    xuqingqing@chinamobile.com
  }
}
}
\maketitle

\begin{abstract}
While deep neural network (DNN)-based video denoising has demonstrated significant performance, deploying state-of-the-art models on edge devices remains challenging due to stringent real-time and energy efficiency requirements. Computing-in-Memory (CIM) chips offer a promising solution by integrating computation within memory cells, enabling rapid matrix-vector multiplication (MVM). However, existing DNN models are often designed without considering CIM architectural constraints, thus limiting their acceleration potential during inference. To address this, we propose a hardware-algorithm co-design framework incorporating two innovations: (1) a CIM-Aware Architecture, CIM-NET, optimized for large receptive field operation and CIM's crossbar-based MVM acceleration; and (2) a pseudo-convolutional operator, CIM-CONV, used within CIM-NET to integrate slide-based processing with fully connected transformations for high-quality feature extraction and reconstruction. This framework significantly reduces the number of MVM operations, improving inference speed on CIM chips while maintaining competitive performance. Experimental results indicate that, compared to the conventional lightweight model FastDVDnet, CIM-NET substantially reduces MVM operations with a slight decrease in denoising performance. With a stride value of 8, CIM-NET reduces MVM operations to 1/77th of the original, while maintaining competitive PSNR (35.11 dB vs. 35.56 dB).
\end{abstract}

\section{Introduction}
Denoising is a critical component of video processing pipelines, enhancing the performance of downstream tasks \cite{21-da2016empirical,22-sadda2018real,23-monakhova2022dancing,25-sadda2018real}. The increasing adoption of automation and intelligent systems has led to the application of denoising algorithms in edge computing scenarios, including autonomous driving, robotics, and satellite remote sensing \cite{24-liu2023lightweight}. These applications impose strict demands on denoising processes with respect to real-time performance and responsiveness. Addressing this challenge necessitates a holistic approach considering both algorithmic design and hardware constraints.

Deep learning advancements have significantly improved the accuracy of video denoising algorithms. Vision transformers (ViT) have garnered attention due to their ability to capture long-range dependencies \cite{33-liang2024vrt}. However, ViT models often exhibit high computational complexity and inference latency. As demonstrated in prior work, ViT-based video denoising models, such as VRT, typically require 25 times more parameters and exhibit 27 times slower inference speeds compared to convolutional denoising models \cite{1-bled2024lightweight}. Consequently, deploying state-of-the-art ViT-based algorithms remains a critical challenge given the limited resources and power constraints of edge devices. In contrast, convolutional neural networks (CNNs) remain effective for video denoising due to their efficiency in local feature extraction and their ability to capture spatial and temporal relationships through specialized convolutional layers \cite{1-bled2024lightweight,2-ding2022scaling}. This makes CNNs applicable in fast video processing tasks and edge computing scenarios.

The advent of Computing-in-Memory (CIM) chips has created new opportunities for accelerating inference, particularly in resource-constrained environments \cite{31-sun2023full}. Unlike traditional von Neumann architectures, which require data transfer between memory and processing units, CIM chips directly integrate matrix-vector multiplication (MVM) computation within memory cells. This architecture reduces the frequency of data transfer from external memory and overcomes the limitations of "memory walls." Furthermore, MVM efficiency in CIM implementations is enhanced by the crossbar nature and physical laws, such as Ohm's Law and Kirchhoff's Law \cite{8-bai2024end}. These characteristics lead to low energy consumption and high computational efficiency for DNN inference.

Numerous efforts have been made to adapt DNN models for CIM chips to improve inference speed \cite{3-kim2022overview,4-yu2021compute}. Considering the immaturity of current CIM chip manufacturing processes, lightweight DNN model methods have been proposed, including quantization \cite{5-jacob2018quantization}, pruning-based sparsity optimization \cite{6-han2015learning}, and knowledge distillation \cite{6-2-gou2021knowledge}. Researchers have also explored allocation strategies for mapping DNN model parameters onto CIM chip arrays, such as the Overlapped Mapping Method (OMM), to further accelerate inference \cite{8-bai2024end,9-zhu2018mixed}. However, these methods often fail to fully leverage the computational advantages of CIM due to a lack of consideration for chip characteristics during model architecture design. This results in suboptimal solutions achieved through post-optimization.

To address this problem, this paper explores DNN model design strategies that fully exploit the computational advantages of MVM operations in CIM chips. We propose a pseudo-convolutional operator, CIM-CONV, which uses a multi-input multi-output structure within each sliding window. In contrast to conventional convolution operations, this operator generates output feature maps of arbitrary sizes from an input patch through a single MVM operation. This approach preserves the spatial continuity of output features and enables flexible configuration of input and output dimensions, tailored to the array structure of the target CIM chip. Based on CIM-CONV, we introduce CIM-NET, a novel video denoising DNN architecture. Building upon the FastDVDnet framework, our model replaces upsampling, downsampling, and smoothing components with CIM-CONV operators. By incorporating a multi-input multi-output processing scheme, CIM-NET substantially reduces the number of MVM operations during inference. Compared to traditional convolution operations involving large kernels and strides, this design preserves spatial coherence more effectively, leading to enhanced denoising performance.

The primary contributions of this paper are as follows:
\begin{itemize}
    \item A hardware-aware operator, CIM-CONV, for efficient feature extraction on CIM platforms. This operator abstracts convolution-like behavior into a single MVM-compatible operation, bridging the gap between deep neural network design and CIM hardware constraints.

    \item A lightweight video denoising architecture, CIM-NET, tailored for MVM-based inference. By restructuring conventional modules in FastDVDnet with CIM-CONV, the proposed CIM-NET achieves substantial reductions in computational cost while maintaining competitive denoising performance.

    \item An empirical study on the trade-off between inference efficiency and denoising performance on CIM hardware. Through extensive evaluation under varying stride settings, we demonstrate that CIM-NET achieves comparable PSNR to FastDVDnet at low strides, while enabling up to 77x reduction in MVM operations with minimal performance degradation (0.45dB) at higher strides.
\end{itemize}

The remainder of this paper is organized as follows: Section II presents background information on video denoising techniques and CIM technology. Section III details the proposed CIM-NET architecture and the design rationale for the CIM-CONV module. Section IV presents experimental validation, including comparative analyses with baseline models, ablation studies, and robustness evaluations. Section V concludes the paper.

\section{Background}
This section provides a brief overview of video denoising techniques, followed by an introduction to CIM technology and its acceleration principles for DNN model inference.

\subsection{Video Denoising Technology}

CNN-based video denoising network models have evolved from shallow architectures, such as MLP-based methods \cite{27-burger2012image}, to deeper networks like DnCNN \cite{28-zhang2017beyond}, which introduced residual learning for Gaussian noise removal. Subsequent innovations, such as FFDNet, employed noise level maps as input conditions to enable blind denoising \cite{29-zhang2018ffdnet}. Furthermore, video denoising extends image-based approaches by exploiting temporal coherence across frames, as seen in DVDnet, which pioneered a two-stage framework combining spatial denoising with optical flow-guided temporal fusion \cite{30-tassano2019dvdnet}. To circumvent inaccuracies in optical flow estimation under heavy noise, FastDVDnet utilized a U-Net backbone with interleaved spatial-temporal convolutions, achieving notable real-time denoising performance through parameter efficiency \cite{26-tassano2020fastdvdnet}. Recent research has also explored deploying lightweight convolutional networks on edge NPU chips. For instance, MFD-Net introduces a spatial attention mechanism and feature downsampling strategy to achieve competitive denoising performance while maintaining low-latency inference on mobile devices \cite{24-liu2023lightweight}. Despite these advancements, most existing models still depend on complex network structures that are computationally intensive and challenging to deploy on low-power platforms. Consequently, designing real-time, low-power video denoising models that can be efficiently deployed on resource-constrained edge devices has become an increasingly important and challenging research area.\\

\subsection{CIM Technology}

\begin{figure}[hbt]
\centering
\includegraphics[width=15pc]{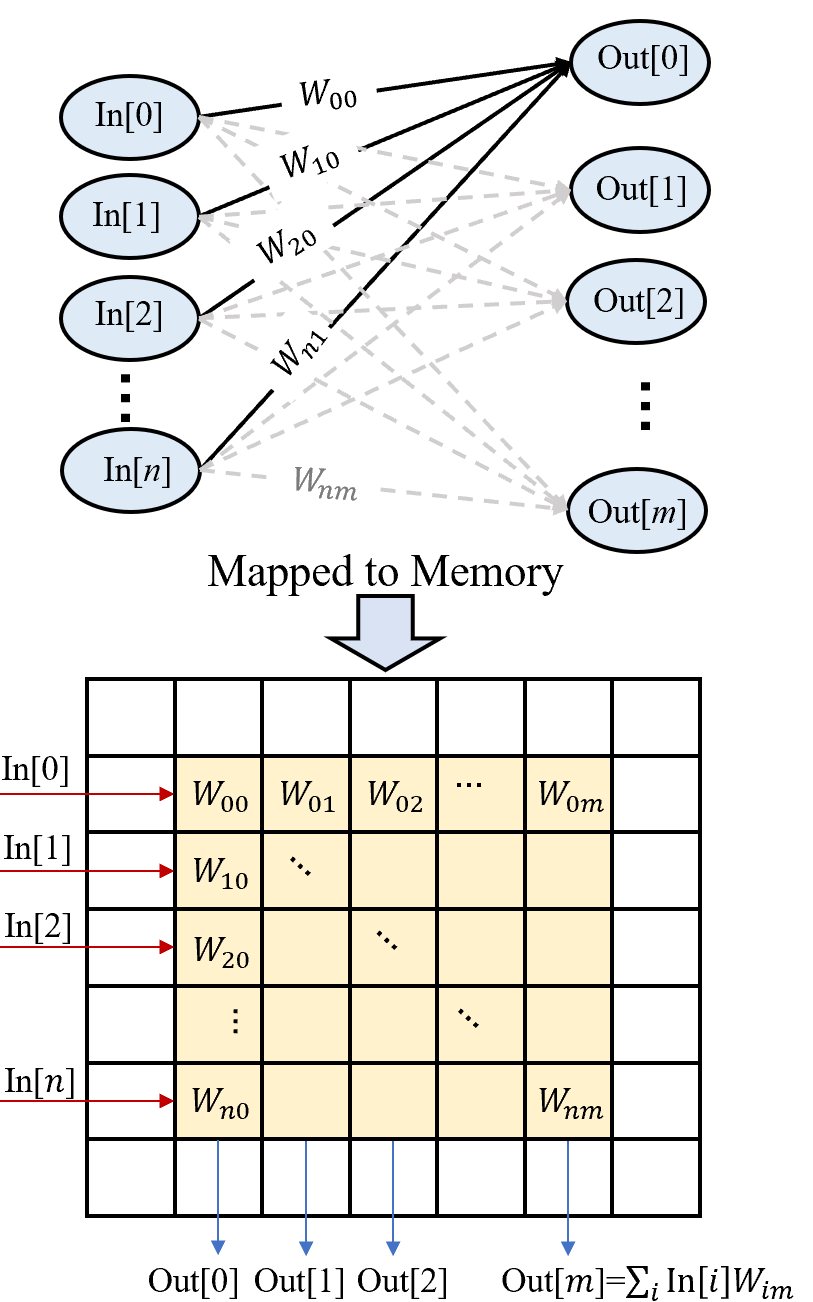}
\caption{MVM operation schematic of CIM chip. A layer of neural metwork is mapped to the memory sub-arrays. Inputs are loaded in parallel as voltage to activate multiple rows, and column currents are summed up based on Kirchhoff’s Law. }
\label{cim_chip}
\end{figure}

In traditional von Neumann architectures, DNN model inference involves numerous MVM operations, resulting in substantial data movement between processing units and memory. This leads to the well-known "power wall" and "memory wall" issues. CIM techniques have been reported to mitigate these bottlenecks, improving energy efficiency and accelerating inference speed. Firstly, computation is enabled directly within storage units, eliminating the frequency of data transfer from and to external memory \cite{31-sun2023full}. Furthermore, MVM operations on CIM architectures are remarkably boosted by the crossbar nature and physical laws, such as Ohm's Law and Kirchhoff's Law \cite{8-bai2024end}. The implementation of a typical MVM operation on a CIM chip is illustrated in Figure \ref{cim_chip} , using an eFlash chip as an example. The weight in a neural network layer is stored as the conductance (denoted by $W_{ij}$ for the $i^{th}$ row and $j^{th}$ column) of the CIM device. The input features are represented as the voltage (denoted as In[i] for ith row). The multiplication between one input feature and one weight is represented by the current through one bit-cell. Hence, the dot product between the input vector and the weight matrix can be performed in the analog domain by accumulating bit-cell currents from the same column. In principle, MVM could be done in a fully parallel fashion if asserting all the rows and all the columns simultaneously.\\

It is noteworthy that when performing convolution operations on CIM chips, a single-channel convolution kernel is typically transformed into a one-dimensional vector and stored in a single row of the CIM array. The complete input features are divided into multiple patches based on the size of the convolution kernel, and are sequentially input into the CIM chip to perform MVM operations. Therefore, the number of MVM operations corresponds to the number of convolutional sliding windows. Given that CIM chips have a fixed MVM operation time, reducing the number of sliding windows can help accelerate the inference speed of network models deployed on CIM chips.

\section{Proposed Model}

This section focuses on optimizing conventional CNN architectures to better align with the computational characteristics of CIM chips, with the objective of reducing inference latency and energy consumption. We begin by conducting a comparative analysis of key factors influencing the inference speed and accuracy of neural network models, followed by the introduction of the proposed architectural design. Lastly, we present CIM-NET, a fast video denoising DNN model tailored for CIM chips.

\subsection{Baseline Model and Optimization Strategy}

\begin{figure}[hbt]
\centering
\includegraphics[width=20pc]{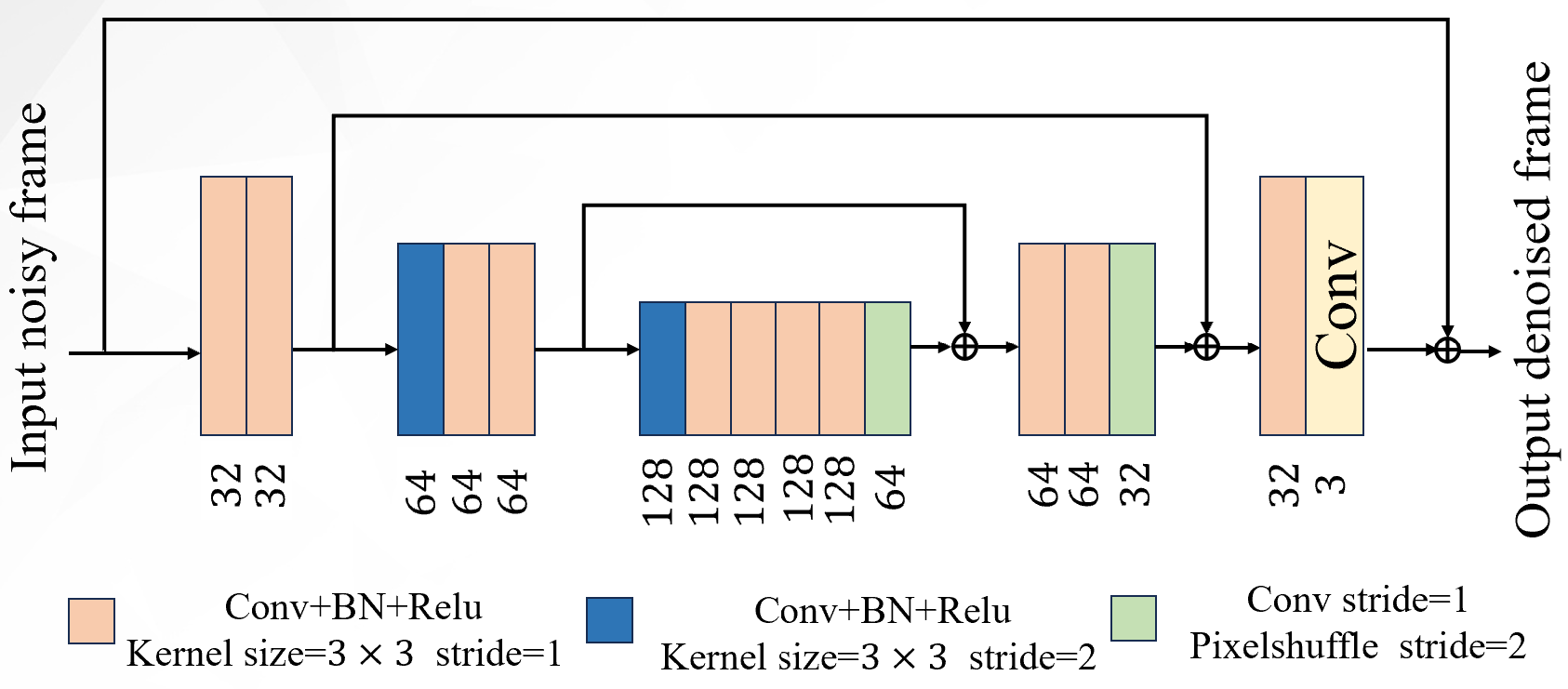}
\caption{Denosing module of FastDVDnet.}
\label{fastdvdnet}
\end{figure}

To ensure the runtime performance and lightweight of the proposed model, we utilize the denoising block in FastDVDnet as the baseline, as shown in Fig.\ref{fastdvdnet}, which contains only well-optimized \(3\times3\) convolutions and \(ReLU\) activation functions. The numbers below each layer of the network in the figure represent the number of convolutional output channels \(C_{out}\). In the \(Conv+Pixelshuffle\) module [20], the output channel of the convolutional layer is set to \(C_{out} \times Stride \times Stride\). This architecture has demonstrated its effectiveness in video denoising task with significantly lower computing times \cite{26-tassano2020fastdvdnet}.

\begin{figure}[hbt]
\centering
\includegraphics[width=22pc]{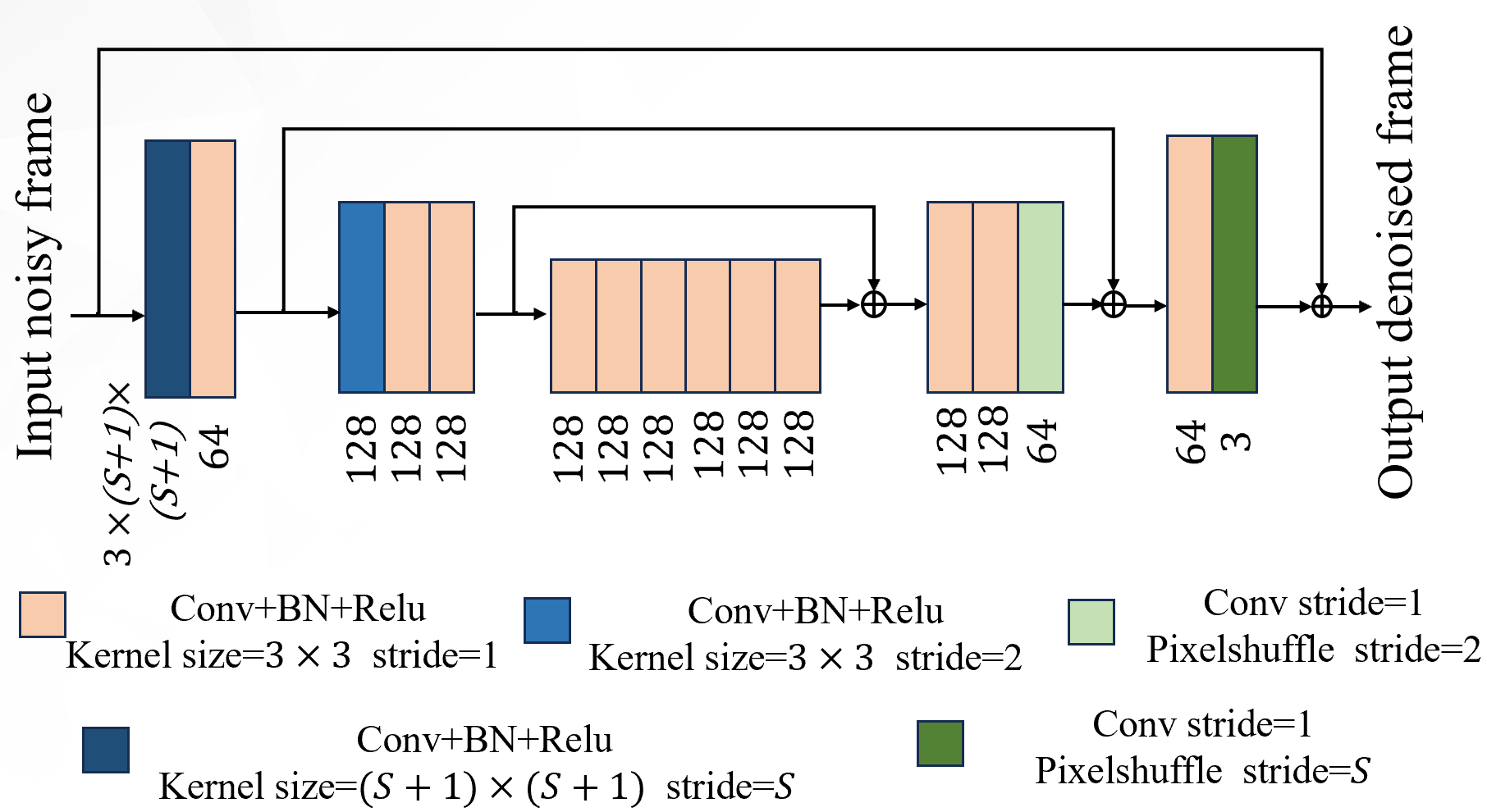}
\caption{Architecture used in Baseline o1.}
\label{o1_net}
\end{figure}

Although the aforementioned architecture exhibits excellent performance on GPU chips, small-sized convolution kernels result in a large number of sliding windows when using CIM chips for convolution operations. To reduce the number of convolutional sliding windows, based on this plain topology, we adopt large-kernel convolution layers with large strides for feature extraction on the original input image. The modified model, denoted as Baseline o1, is illustrated in Figure \ref{o1_net}. \(S\) represents the stride value, and \(S+1\) represents the size of the convolution kernel. At the end of the network is a \(Conv+Pixelshuffle\) for reconstruction. The number of output channels in the convolutional layer before the last \(Pixelshuffle\) is set to \(3 \times (S+1)\times (S+1)\).

\begin{figure}[hbt]
\centering
\includegraphics[width=21pc]{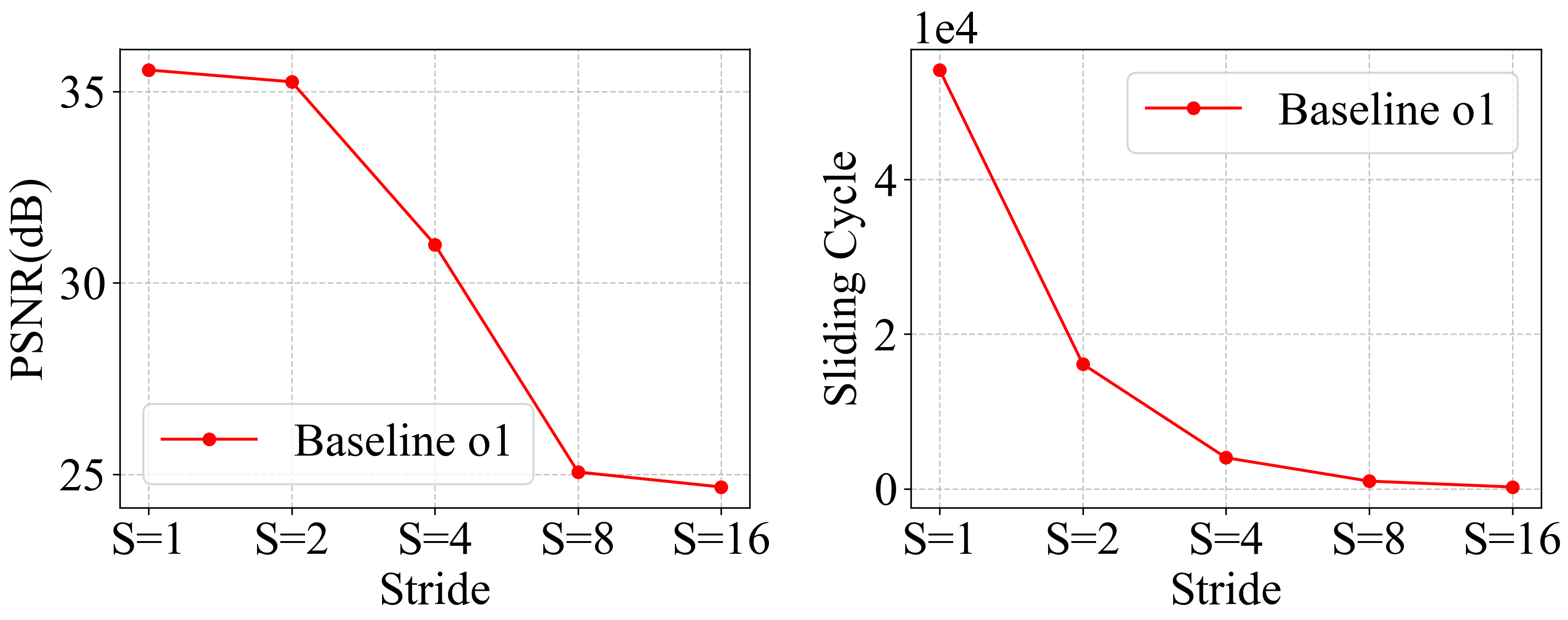}
\caption{Baseline o1 performance varies with stride value.}
\label{o1_result}
\end{figure}

With the input image size set to \(96 \times 96\) and the number of input channels fixed at 3, Figure \ref{o1_result} illustrates the variation in the number of sliding windows and the corresponding denoising performance under different values of \(S\). The input noisy videos have a PSNR of 24.61 dB, and detailed experimental settings are provided in Section IV. The result of \(S=1\) corresponds to the original model configuration shown in Fig.\ref{fastdvdnet}. It can be observed that increasing the stride and kernel size of the first convolutional layer significantly reduces the number of sliding windows, thereby accelerating inference on the CIM chip. Specifically, the inference speed improves proportionally with \(S^2\). However, this gain in efficiency comes at the cost of denoising quality. As \(S\) increases, the model exhibits a notable decline in PSNR performance, particularly when \(S>4\), indicating a clear trade-off between computational speed and denoising effectiveness.

\begin{figure}[hbt]
\centering
\includegraphics[width=20pc]{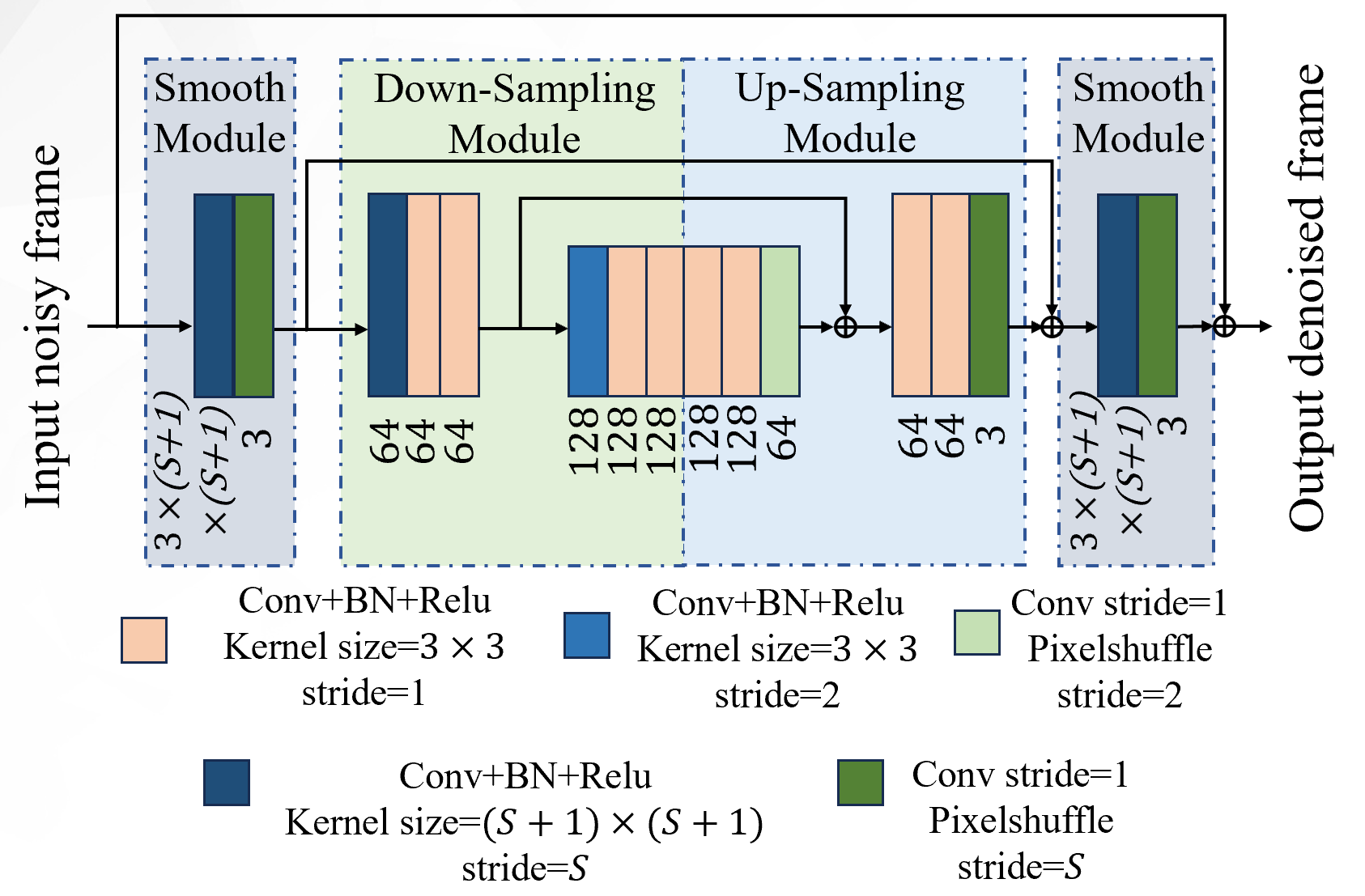}
\caption{Architecture used in Baseline o2.}
\label{o2_net}
\end{figure}

\begin{figure}[hbt]
\centering
\includegraphics[width=21pc]{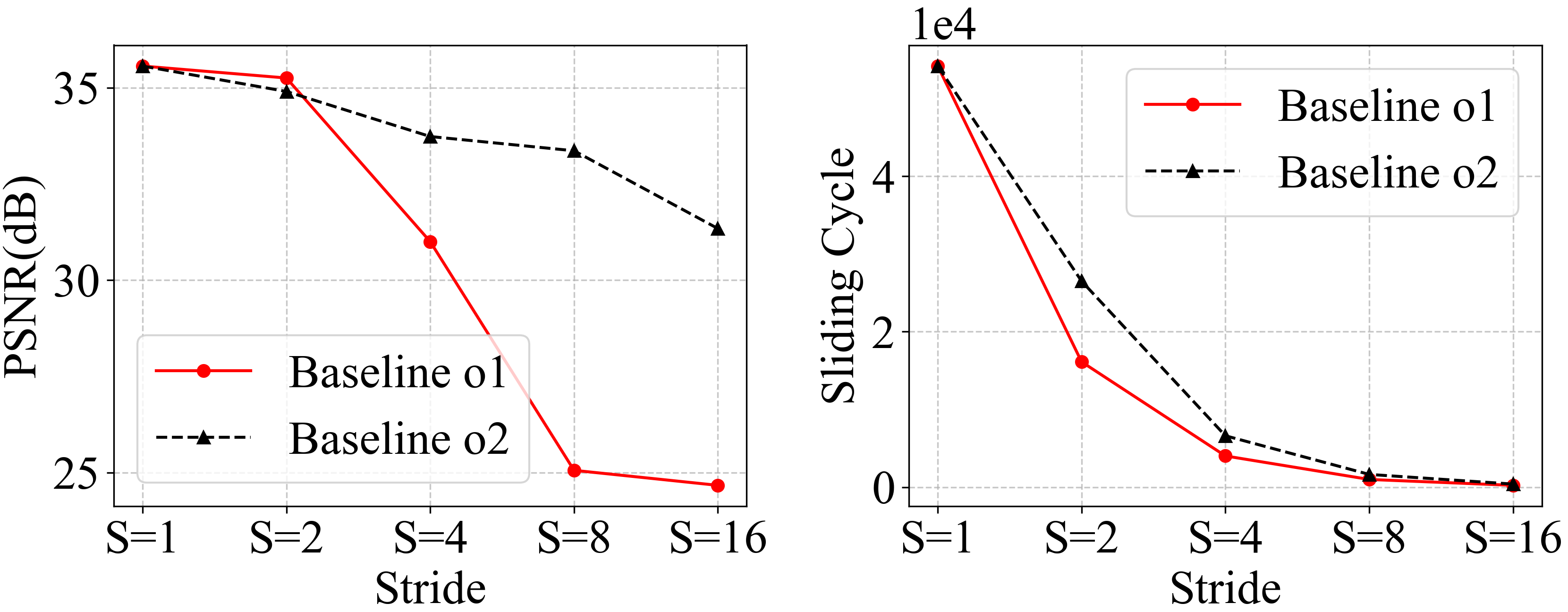}
\caption{Baseline o2 performance varies with stride value.}
\label{o2_result}
\end{figure}

To mitigate the performance degradation caused by increasing the convolutional kernel size, we introduce smoothing modules before the downsampling process and after the upsampling process. These modules are designed to alleviate block artifacts introduced during the reconstruction stage, and the resulting model is referred to as Baseline o2. Each smoothing module adopts a U-Net architecture comprising a cascaded large stride convolution layer and a \(Pixelshuffle\) layer. The overall model structure is illustrated in Figure \ref{o2_net}, and the comparative results are shown in Figure \ref{o2_result}.

By adding smoothing modules, the model achieves a notable improvement in denoising performance with only a marginal increase in the total number of sliding windows, particularly when \(S\)  is greater than 4. It is worth noting that, when \(S=2\), the denoising performance of Baseline o2 is slightly inferior to Baseline o1. This may be attributed to the effect of the \(Pixelshuffle\) operation, which can disrupt the spatial continuity of convolutional outputs and hinder subsequent feature extraction. When the stride value is small, this limitation becomes the dominant factor affecting model performance, ultimately leading to Baseline o2 underperforming Baseline o1 in such cases.

\subsection{CIM-NET}

\begin{figure}[hbt]
\centering
\includegraphics[width=20pc]{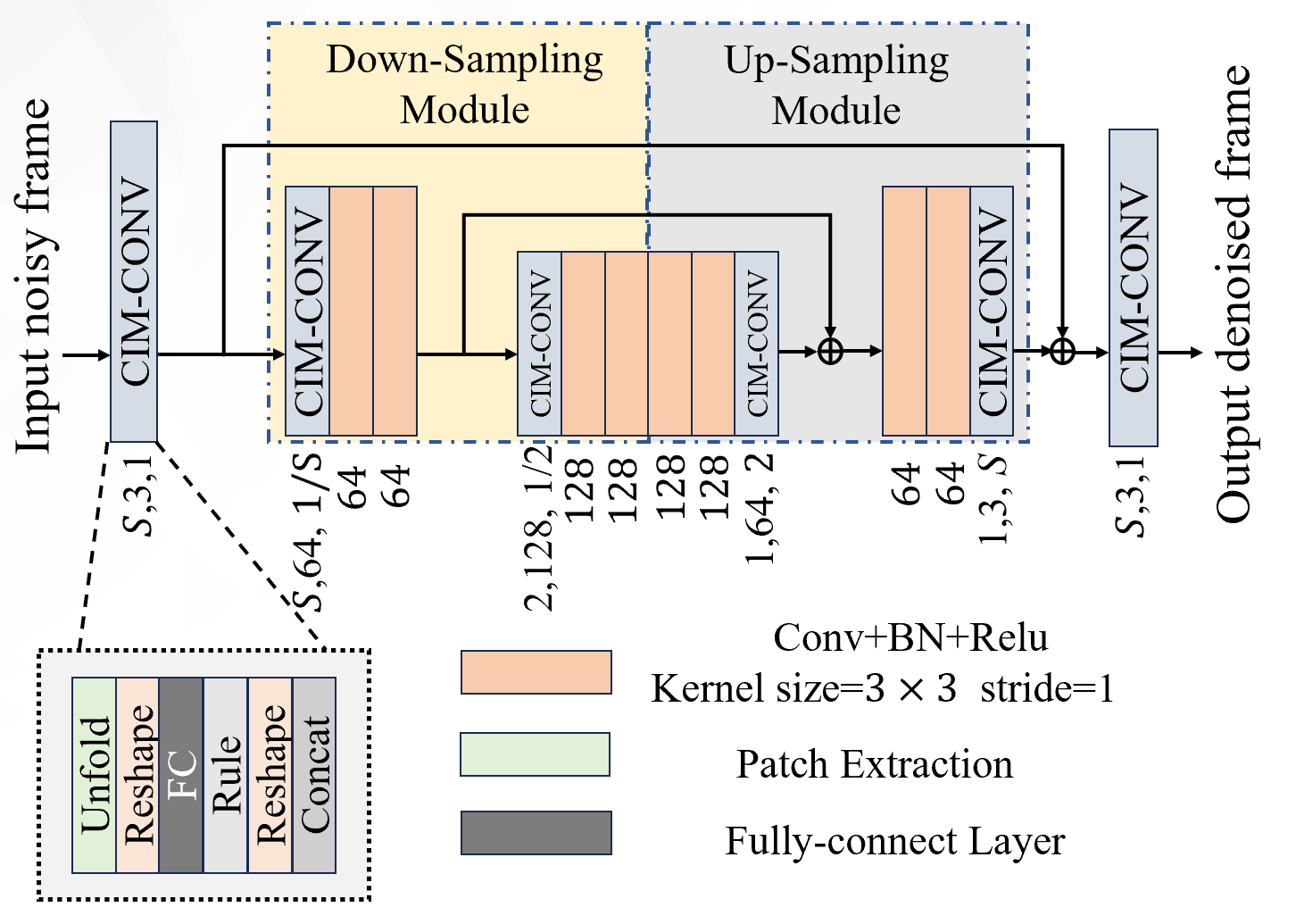}
\caption{Architecture used in CIM-NET.}
\label{CIM-NET}
\end{figure}

As demonstrated in the preceding comparative analysis, expanding the receptive field in the initial convolutional layer effectively reduces the number of sliding cycles, while the incorporation of upsampling-downsampling smoothing modules compensates for the precision loss introduced by aggressive downsampling and reconstruction. Motivated by these insights, we propose CIM-NET — a novel video denoising network specifically optimized for fast inference on CIM processors (Fig.\ref{CIM-NET}). Unlike the aforementioned large step convolution structures, a CIM-CONV module is proposed and used here to directly obtain the output feature map of any size from the receptive field of the input feature. This process can be regarded as a pseudo-convolution operation capable of performing smoothing, upsampling, and downsampling within denoising models. A detailed description of this module is provided in following sections.

The input image first undergoes feature mapping through the CIM-CONV module, which does not reduce the dimensionality of the input features. The three parameters under each CIM-CONV layer represent the stride value \(S\), the output feature layer \(C_{out}\), and scaling factor \(F_{scale}\). The output of this module is downsampled through CIM-CONV, followed by multi-layer \(3 \times 3\) convolution for high-dimensional feature extraction. After dual downsampling steps, two-stage upsampling using convolutional layers coupled with CIM-CONV is conducted. Lastly, the reconstructed image is smoothed using another CIM-CONV. Empirical observation reveals superior denoising performance when directly estimating original images through CIM-CONV compared to conventional noise component estimation. Consequently, we eliminate residual connections in favor of direct output generation.
\begin{figure}[hbt]
\centering
\includegraphics[width=20pc]{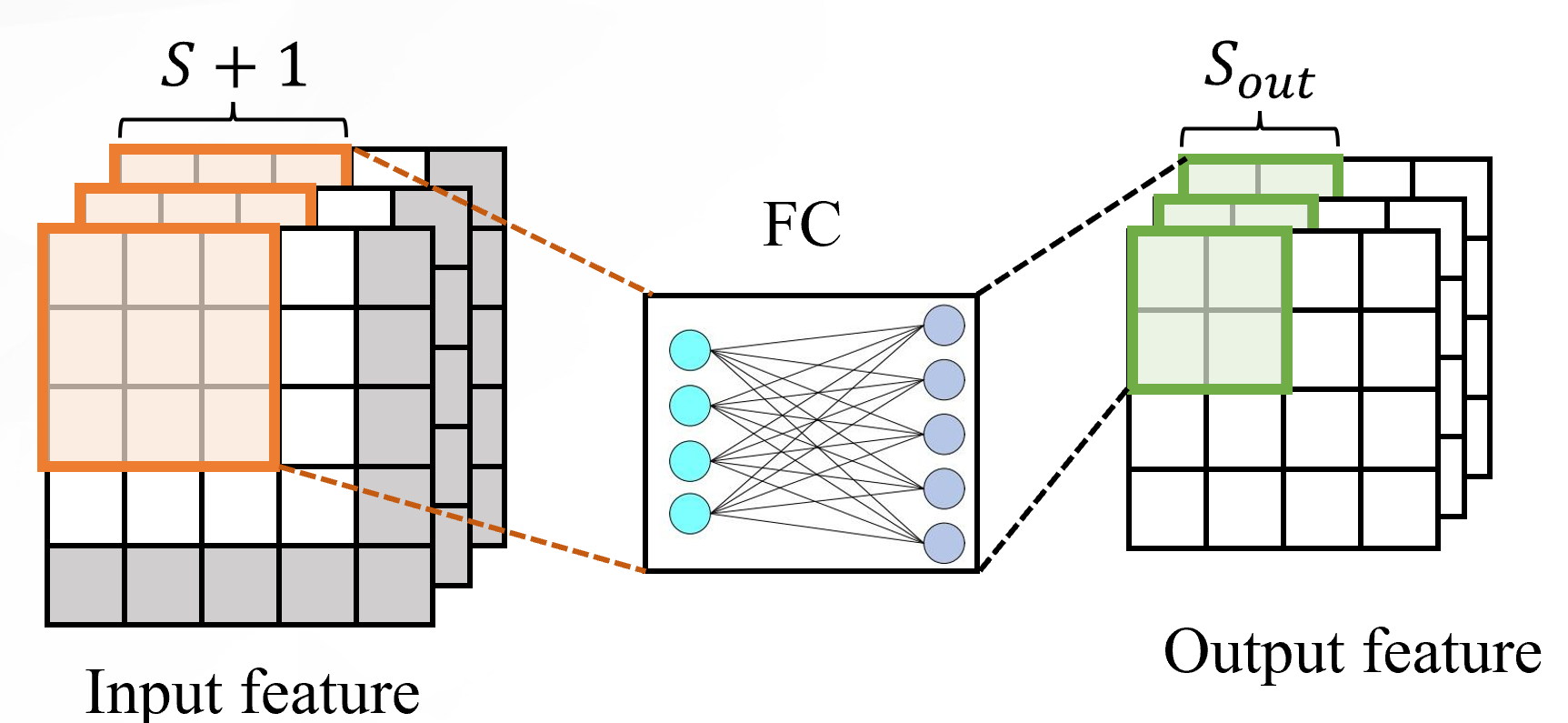}
\caption{The schematic diagram of CIM-CONV.}
\label{CIM-CONV}
\end{figure}

The proposed CIM-CONV module implements a learnable upsampling operator that combines patch-based processing with fully connected transformations, serving as an alternative to conventional convolution/transposed-convolution operations. The key components and operational flow are described as follows:

\textit{Patch Extraction}: Given input tensor \(x \in R^{N\times C\times H\times W}\), after performing \(Padding=1\) operation on the input features, we utilize tensor unfolding (Unfold) to decompose input features into overlapping patches of size \((S+1)\times(S+1)\) with stride \(S\). The Reshape operation expands the multi-channel grouping features\(C_{in}\times(S+1)\times(S+1)\) into one-dimensional vectors.

\textit{FC Transformation}: The fully-connected (FC) operation maps each expanded one-dimensional vector to the output with the size \(C_{out}\times S_{out}\times S_{out}\), where \(S_{out}=S\cdot F_{scale}\). The scaling factor \(F_{scale}\) of CIM-CONV module determines the change in feature dimension. This transformation is followed by \(Rule\) as the activation function.

\textit{Spatial Reconstruction}: After fully connected mapping, use Reshape operation to transform the one-dimensional output result into a three-dimensional matrix, and finally concatenate the output features of each group to obtain the complete output result.

Compared to the \(Conv+Pixelshuffle\) scheme used for upsampling and feature smoothing in Baseline o2, the proposed CIM-CONV maintains a small number of sliding windows while eliminating the channel-to-spatial conversion process, thereby preserving the spatial continuity of the output features. Moreover, by adjusting the fractional or integer scaling factor \(F_{scale}\), CIM-CONV can flexibly achieve arbitrary spatial scaling of the input features. It is also worth noting that, under identical stride conditions, the CIM-CONV module replaces the dual-layer convolution structure employed by the smoothing module in Baseline o2 with a sliding fully connected layer. This substitution results in a 17.39\% reduction in sliding cycles, thereby reducing the total number of MVM operations during inference on CIM chips.

\section{Experiments and Discussion}

This section evaluates the robustness of the proposed model by comparing its denoising performance under various noise conditions. Furthermore, ablation studies are conducted to validate the effectiveness of the proposed CIM-CONV module.

\subsection{Experimental Setup}

The training and testing dataset is sourced from the DAVIS dataset \cite{32-KhoRohrSch_ACCV2018}, and a total of 256000 training samples are randomly cropped with a patch size of \(96\times96\). 80\% of it is used for training, 10\% for validation, and 10\% for testing. The noisy frames are generated by adding Additive White Gaussian Noise (AWGN) of \(\sigma\in[5,50]\) to clean patches. \(\sigma\) represents the standard deviation of noise. The training process is implemented in Pytorch \cite{18-paszke2017automatic}. The Mean Squared Error (MSE) loss function is used to minimize the discrepancy between the model’s output and the clean target frames. The ADAM algorithm is applied to minimize the loss function \cite{19-kingma2014adam}.  The number of epochs for training is set to 100, and the mini-batch size is 96. The scheduling of the learning rate is the same as the setting in \cite{26-tassano2020fastdvdnet}, which is initially set to \(1e^{-3}\) for the first 50 epochs, then changes to \(1e^{-4}\) for the following 10 epochs, and finally switches to \(1e^{-6}\) for the remaining of the training.

\subsection{Performance Validation of CIM-NET}

\begin{figure}[hbt]
\centering
\includegraphics[width=20pc]{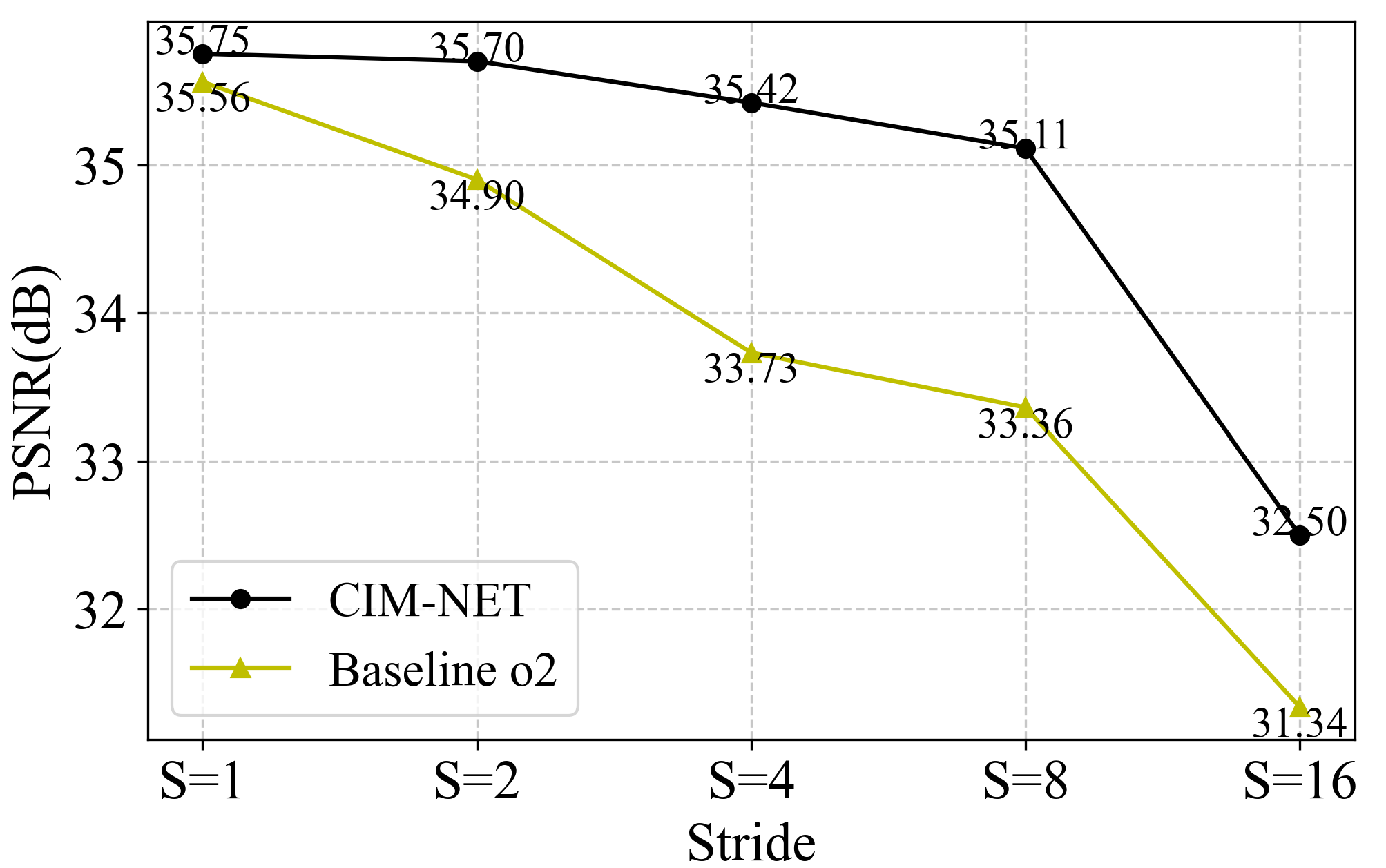}
\caption{CIM-NET performance varies with stride value.}
\label{diff_strides}
\end{figure}

To validate the superiority of the proposed model in CIM-accelerated inference tasks, we compared the denoising performance between CIM-NET and the baseline o2 model under different stride configurations, and the results are illustrated in Fig.\ref{diff_strides}. The experiments were conducted with an input noise standard deviation \(\sigma=15\), achieving an average input video PSNR of 24.61 dB.

When \(S=1\), CIM-NET adopts a stride of 1 in the smoothing module and stride 2 in the downsampling module, both maintaining a \(3\times 3\) receptive field. This configuration aligns with the denoising module settings in FastDVDNet, ensuring equivalent sliding window operations to baseline o2. Under these conditions, CIM-NET achieves a PSNR of 35.75 dB, marginally outperforming the baseline system's 35.56 dB. This demonstrates that the SIM-CONV module can enhance network performance through optimized upsampling processes even under small stride conditions. As the stride increases, the required sliding window operations for CIM-NET inference decrease by a factor of 1/\(S^2\), accompanied by a gradual performance degradation. Nevertheless, CIM-NET maintains significantly superior denoising results compared to baseline o2. Notably, when the stride expands to 8, the sliding window operations on CIM chips are reduced to 1/64 of the original while the PSNR degradation remains limited to 0.64 dB. Compared with the denoising module settings in FastDVDNet, this setting reduces the number of MVM operation by about 1/77 of the original (CIM-CONV reduces the number of layers), while the PSNR degradation remains limited to 0.45 dB. These findings indicate that our proposed architecture effectively improves feature extraction accuracy for large receptive fields and high-stride configurations. Compared with conventional networks, it demonstrates enhanced comprehensive performance on CIM chips, achieving an optimal balance between computational efficiency and denoising quality.
\begin{figure}[hbt]
\centering
\includegraphics[width=20pc]{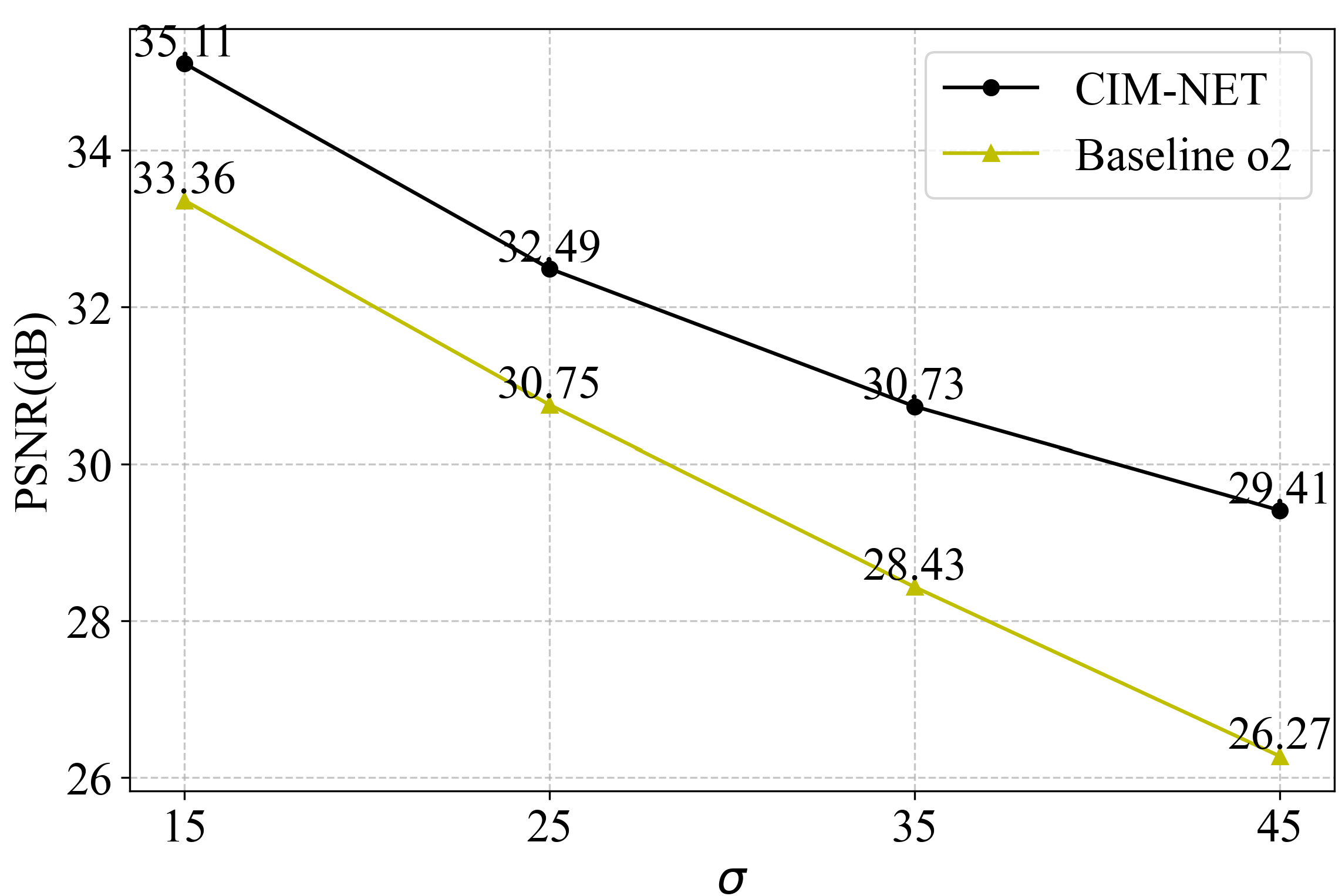}
\caption{CIM-NET performance varies under different SNR scenarios.}
\label{diff_noise}
\end{figure}

Fig.\ref{diff_noise} illustrates the variation in denoising performance of the models under different signal-to-noise ratio (SNR) conditions, with both models configured with a stride of 8. Overall, CIM-NET achieves a denoising performance improvement of 1.74 dB to 2.14 dB across varying SNR levels compared to Baseline o2. As the noise intensity increases, CIM-NET demonstrates relatively higher robustness. This result suggests that the multi-output operations in CIM-NET facilitate superior denoising effects by leveraging the spatial continuity of input features.

\subsection{Ablation Study}
\begin{figure}[hbt]
\centering
\includegraphics[width=20pc]{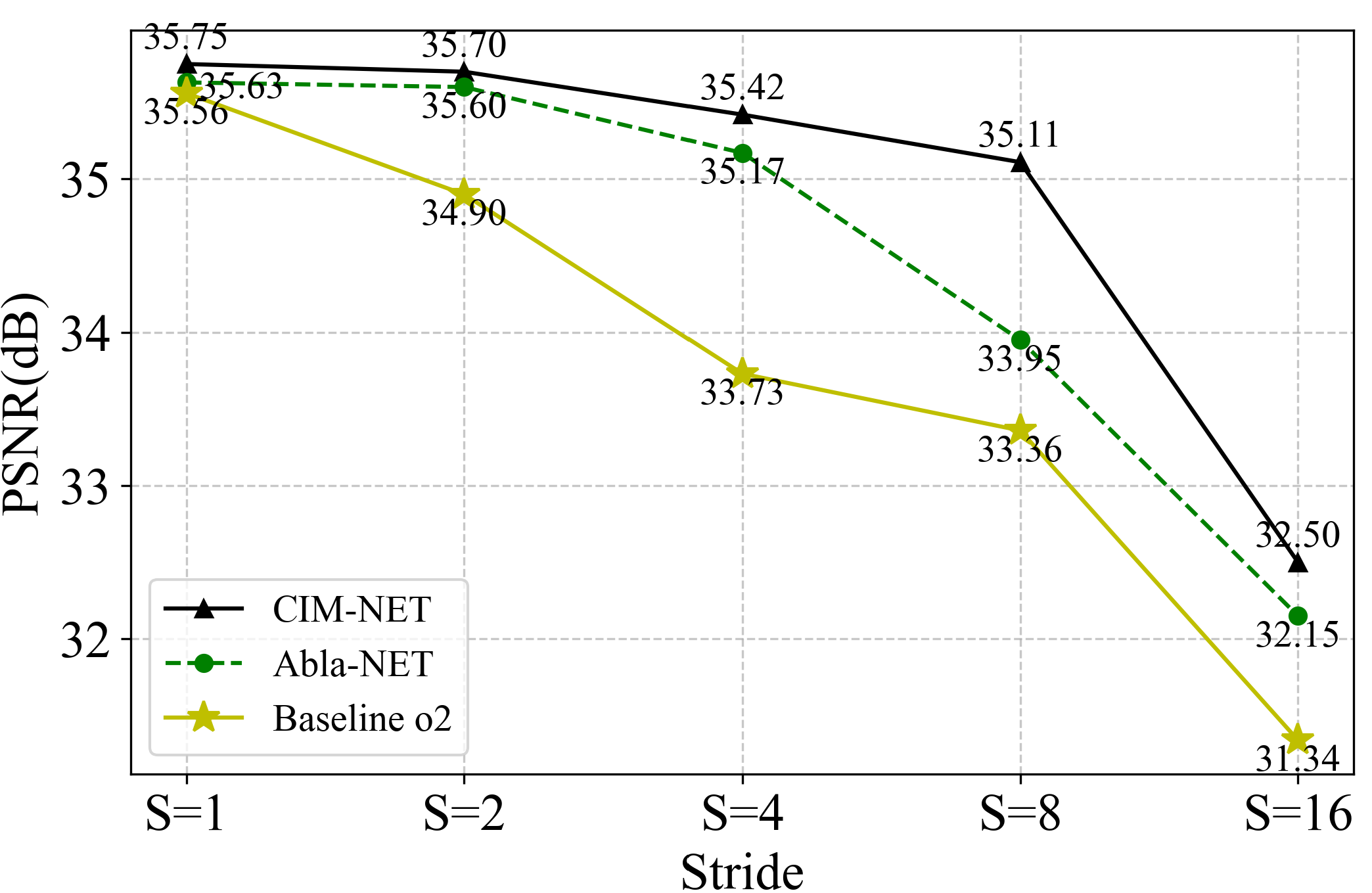}
\caption{Evaluation of denoising performance with partial deployment of CIM-CONV operator.}
\label{abla_result}
\end{figure}

To validate the contributions of CIM-CONV module in difference process, we construct Abla-NET - a variant that retains Baseline o2's original upsampling and downsampling architecture while exclusively substituting its smoothing modules with CIM-CONV. The denoising outcomes are presented in Fig.\ref{abla_result}. Comparative analysis between the baseline and Abla-NET reveals that merely replacing the smoothing modules yields significant SNR improvements (up to 1.44 dB when S=4). When extending CIM-CONV implementation to upsampling and downsampling modules in the full CIM-NET, we observe additional performance gains (0.25 dB when S=4) without increasing network parameter count. The above results indicate that compared with the traditional models, the CIM-CONV module can achieve more accurate feature extraction under large stride sizes and receptive fields, as well as more accurate feature reconstruction during the upsampling process. Simultaneously using the CIM-CONV operator in the above process can reduce the number of MVM operations on CIM chips while maintaining optimal denoising performance.

\section{Conclusion}
This work presents CIM-NET, a hardware-aware video denoising architecture optimized for CIM acceleration. Through systematic analysis of MVM operation patterns in CIM chips, we identify that conventional CNN architectures suffer from excessive sliding window computation problem. To address these limitations, the proposed CIM-CONV module integrates patch decomposition, cross-scale feature transformation, and learnable spatial reconstruction into a unified operator, significantly reducing the required MVM operations while maintaining excellent denoising performance. Experimental validation demonstrates that the CIM-NET reduces the number of MVM operation by about 1/77 of the original compared with traditional FastDVDnet (at stride=8), while maintaining competitive PSNR (35.11 dB vs 35.56 dB). Across SNR conditions, the proposed model also maintains about 2dB superiority over the Baseline O2 model, demonstrating its robustness and superiority in noisy environments. These results underscore the importance of co-designing neural operators with underlying hardware constraints and provide new ideas for developing next-generation edge intelligence systems.
\bibliography{cim_net.bib}
\bibliographystyle{cim_net.bst}

\end{document}